\documentclass[a4paper, 10pt, conference]{ieeeconf}

\IEEEoverridecommandlockouts
\usepackage[utf8]{inputenc}
\usepackage{amssymb}
\usepackage{amsmath}
\usepackage{upgreek}
\DeclareMathOperator*{\argmax}{arg\,max}
\DeclareMathOperator*{\argmin}{arg\,min}
\usepackage{graphicx}
\usepackage{multirow}
\usepackage{caption, subcaption}
\usepackage{tikz}
\usepackage{tikz-3dplot}
\usetikzlibrary{bayesnet}
\usepackage[ruled, linesnumbered]{algorithm2e}
\usepackage{url} 
\usepackage{mathtools}

\newcommand{\pose}[1]{$\mathbf{T}_{\text{#1}}$}

\newcommand{\coordinatesystem}[1]{$\underrightarrow{\mathcal{F}}_{\text{#1}}$}

\title{\LARGE \bf
Simulation-based Bayesian inference for robotic grasping
}

\author{Norman Marlier$^{1}$ 
Olivier Brüls$^{2}$ %
Gilles Louppe$^{3}$ %
\thanks{*The authors come from the Univeristy of Liège, Belgium}
\thanks{$^{1}${\tt\small norman.marlier@uliege.be}}%
\thanks{$^{2}${\tt\small o.bruls@uliege.be}}%
\thanks{$^{3}${\tt\small g.louppe@uliege.be}}%
}

\begin{document}

\maketitle
\thispagestyle{empty}
\pagestyle{empty}

\begin{abstract}
General robotic grippers are challenging to control because of their rich nonsmooth contact dynamics and the many sources of uncertainties due to the environment or sensor noise. 
In this work, we demonstrate how to compute 6-DoF grasp poses using simulation-based Bayesian inference through the full stochastic forward simulation of the robot in its environment while robustly accounting for many of the uncertainties in the system. A Riemannian manifold optimization procedure preserving the nonlinearity of the rotation space is used to compute the maximum a posteriori grasp pose. Simulation and physical benchmarks show the promising high success rate of the approach.

\end{abstract}

\section{Introduction}
Industrial grasping works very well in highly structured environments with few uncertainties. However, complex applications requiring great flexibility have recently gained a lot of interest. For such tasks, dealing with uncertainties becomes key to robust performance.

While previous methods relied on simplified surrogates of the likelihood function, we bring a novel simulation-based approach for full Bayesian inference based on a deep neural network surrogate of the likelihood-to-evidence ratio. By framing robotic grasping as an inference task, we demonstrate the general applicability of simulation-based inference algorithms to complex robotic tasks and their usefulness to deal with uncertainties.

We summarize our contributions as follow:
\begin{itemize}
    \item We bring simulation-based Bayesian inference methods~\cite{cranmer2020frontier} to robotic grasping.
    \item We make use of Riemannian manifold optimization to deal with the nonlinearity of the rotation space.
    \item We validate our method on simulated and real experiments. Results show promising grasping performances.
\end{itemize}

\section{Problem statement}
\begin{figure}
\centering
    \resizebox{\linewidth}{!}{
        \includegraphics[height=5cm]{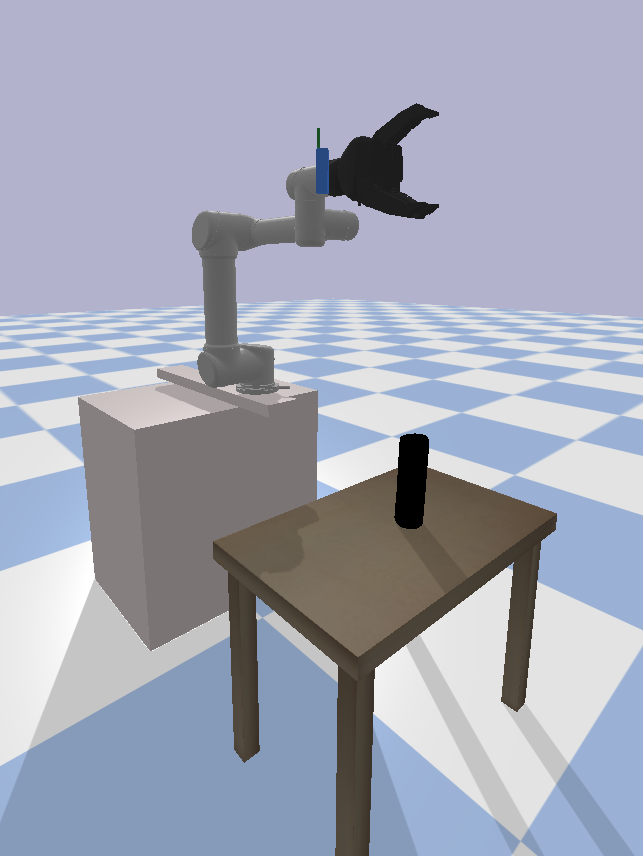}
        \enspace
        \includegraphics[height=5cm]{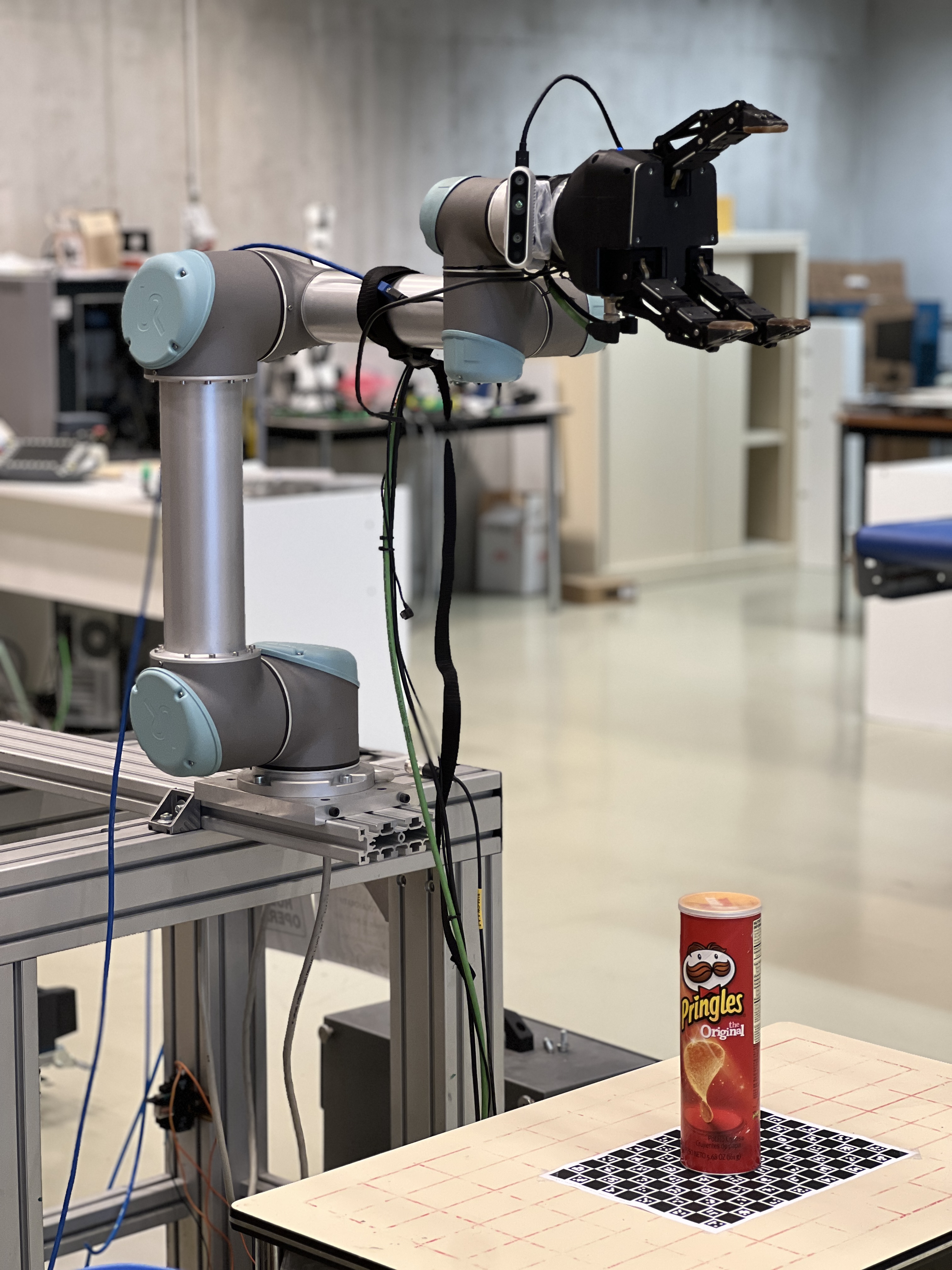}
    }
    \caption{Our benchmark scene. (left) The simulated environment. (right) The real setup.}
    \label{fig:scene}

\end{figure}

We consider the problem of planning 6-DoF hand configurations of a general robotic gripper for unknown rigid objects placed on a table and observed through multi-view depth images (Fig.~\ref{fig:scene}).

\subsection{Description}
The robot arm (6 or 7 DoF) evolves in a cubic workspace with a planar tabletop. It is equipped with a robotic gripper and observes the scene with a depth camera mounted on its flange. Depth images, captured along a predefined trajectory, are fused into a Truncated Signed Distance Function (TSDF) voxel grid~\cite{curless1996volumetric}. Then, we search for the most plausible hand configuration given a successful grasp and the TSDF voxel grid. Finally, a joint trajectory is computed by a path planner based on the TSDF to reach the hand pose and grasp the object in order to remove it from the table.

\subsection{Notations}
\textbf{Frames} We use several reference frames in our work. The world frame \coordinatesystem{W} and the workspace frame \coordinatesystem{S} can be choosen freely and are not tied to a physical location. \coordinatesystem{B}, \coordinatesystem{C}, \coordinatesystem{F}, \coordinatesystem{E} correspond respectively to the robot base, the camera, the flange and the tool center point (TCP).
\par
\textbf{Hand configuration} The hand configuration $\mathbf{h} \in \mathcal{H} = \mathbb{R}^{3}\times \text{SO(3)}$ is defined as the combination of the pose \pose{SE}$ = (_{\text{S}}\mathbf{t}_{\text{SE}},\mathbf{R}_{\text{SE}}) \in \mathbb{R}^{3}\times \text{SO(3)}$ of the hand, where $_{\text{S}}\mathbf{t}_{\text{SE}}$ is the vector $\Vec{SE}$ expressed in \coordinatesystem{S}. We parametrize the rotation $\mathbf{R}_{\text{SE}}$ with quaternions.
\par
\textbf{Binary metric} A binary variable $S \in \{0, 1\}$ indicates if the grasp fails ($S=0$) or succeeds ($S=1$).
\par
\textbf{Observation} Given the depth images $\mathcal{I}_{k} = \{I_{0},...,I_{k}\}$ with their corresponding transformations camera to world $\Gamma_{k} = \{\mathbf{T}^{0}_{\text{WC}},..., \mathbf{T}^{k}_{\text{WC}}\}$ and camera intrinsic matrix $K$, we construct a TSDF voxel grid $\mathbf{V}$ with $N^{3}$ voxels, representing the workspace of size $l$.
\par
\textbf{Latent variables} Unobserved variables $\mathbf{z}$ capture uncertainties about the nonsmooth dynamics of contact, the sensor noise, as well as the geometry of the object (see Section.\ref{sec:data-generation}).

\subsection{Probabilistic modeling}
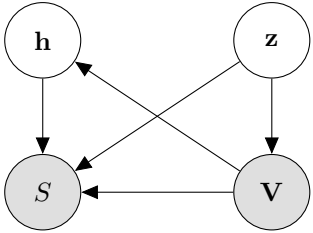
\begin{figure}[h]
    \centering
    \begin{tikzpicture}[scale=1]
 \node[latent, minimum size=1cm] (h) {$\mathbf{h}$};
 
 \node[latent, right=2cm of h, minimum size=1cm] (z) {$\mathbf{z}$};
 
 \node[obs, below=1cm of z, minimum size=1cm]  (V) {$\mathbf{V}$};
 
  \node[obs, below=1cm of h, minimum size=1cm] (S) {$S$};
  
  \edge{h, z}{S};
  \edge{z}{V};
  \edge{V}{h};
  \edge{V}{S};
 

  


  

  
  
\end{tikzpicture}
    \caption{Probabilistic graphical model of the environment. Gray nodes correspond to observed variables and white nodes to unobserved variables.}
    \label{fig:graphical_model}
\end{figure}

We model the scene and the grasping task according to the Bayesian network shown in Fig.~\ref{fig:graphical_model}.
The variables $S, \mathbf{V}$ and $\mathbf{h}$ are modelled as random variables to capture the noise in sensors, uncertainties in the dynamics, as well as our prior beliefs about the hand configuration. The structure of the Bayesian network is motivated by the fact that $S$ is dependent on $\mathbf{h}$, $\mathbf{V}$ and $\mathbf{z}$, $\mathbf{h}$ is dependent of $\mathbf{V}$  and $\mathbf{V}$ is dependent on $\mathbf{z}$. This structure also enables a direct and straightforward way to generate data: $\mathbf{h}$ and $\mathbf{z}$ are sampled from their respective prior distributions while $S$ and $\mathbf{V}$ can be generated using forward physical simulators.

\subsection{Objectives}
Given our probabilistic graphical model, we formulate the problem of grasping as the Bayesian inference of the hand configuration $\mathbf{h}^{*}$ that is a posteriori the most likely given a successful grasp and a TSDF voxel grid $\mathbf{V}$. That is, we are seeking for the maximum a posteriori (MAP) estimate
\begin{equation}
\label{eq:map}
\mathbf{h}^{*} = \argmax_{\mathbf{h}}~p(\mathbf{h}|S=1, \mathbf{V}),
\end{equation}
from which we then compute the joint trajectory
\begin{equation}
    \tau_{1:m} = \Lambda(\tau_{0}, \textsc{ik}(\mathbf{h}^{*}), \mathbf{V})
\end{equation}
where $\textsc{ik}$ is an inverse kinematic solver, $\tau_{1:m}$ are waypoints in the joint space, $\tau_{m} = \textsc{IK}(\mathbf{h}^{*})$ and $\Lambda$ is a path planner.

\section{Related work}
Probabilistic approaches for grasping problems are usually based on likelihood functions which model the probability of success or a grasp quality metric with respect to an observation and a grasp pose. Then, different methods can be used to find the maximum likelihood estimate (MLE) which corresponds to the final grasp pose. Numerical optimization can be used when the likelihood is modeled by differentiable models~\cite{lu2020planning}. Direct regression of the MLE with a learnt model generates quick output but without capturing the full distribution~\cite{cai2022real}. Other approaches identify the maximum likelihood estimate based on a list of candidates computed through a grasp map on the sensor space~\cite{breyer2020volumetric, article}. Similar to our work, \cite{van2020learning} learn models respectively for the likelihood and the prior. Then, they can optimize via gradient descent the posterior density. Contrary to our work, they use Euler angles which can lead to gimbal lock and singularities. Our method preserves the topology by using Riemannian gradient descent.

From a statistical perspective, several Bayesian likelihood-free inference algorithms~\cite{marin2012approximate, beaumont2002approximate, Papamakarios2019SequentialNL, SNPEA, SNPEB, APT, pmlr-v119-hermans20a} have been developed to carry out inference when the likelihood function is implicit and intractable. 
These methods operate by approximating the posterior through rejection sampling or by learning parts of the Bayes' rule, such as the likelihood function, the likelihood-to-evidence ratio, or the posterior itself. 
These algorithms have been used across a wide range of scientific disciplines such as particle physics, neuroscience, biology, or cosmology~\cite{cranmer2020frontier}.
To the best of our knowledge, our work is one of the first to apply one of those for the direct planning successful grasps.
More specifically, we rely here on amortized neural ratio estimation~\cite{pmlr-v119-hermans20a} to carry out inference within seconds for any new observation $\mathbf{V}$. In contrast, an approach such as ABC~\cite{marin2012approximate, beaumont2002approximate} could take up to hours to determine a single hand configuration $\mathbf{h}$ since data would need to be simulated on-the-fly for each observation $\mathbf{V}$ due to the lack of amortization of ABC. 
Neural posterior estimation~\cite{APT} is also amortizable but would have required new methodological developments to be applicable on distributions defined on manifolds, such as those needed here for the rotational part of the pose.

\section{Likelihood-free Bayesian inference for multi-fingered grasping}
\label{sec:method}

From the Bayes's rule, the posterior of the hand configuration is
\begin{equation}
\label{eq:proba_cond}
\begin{split}
p(\mathbf{h}|S, \mathbf{V})  = \frac{p(S \mid \mathbf{h}, \mathbf{V})}{p(S\mid \mathbf{V})}p(\mathbf{h}\mid \mathbf{V}). \\
\end{split}
\end{equation}
\subsection{Priors}
\textbf{Position} The prior over the position $_{\text{S}}\mathbf{t}_{\text{SE}} \coloneqq \mathbf{x}_{\text{E}}$ is a uniform distribution over all the dimensions. We first use a uniform distribution over the cube of length $[-1, 1]^{3}$, called $p(\mathbf{u})$ and then use the bijection $\text{B}(\mathbf{u}; \mathbf{V}):[-1, 1]^{3} \rightarrow [x_{\text{low}}, x_{\text{high}}] \times [y_{\text{low}}, y_{\text{high}}] \times [z_{\text{low}}, z_{\text{high}}]$ to compute $\mathbf{x}_{\text{E}}$, where the bounds are chosen to be the dimensions of the object voxel axis aligned bounding box. Then, $p(\mathbf{x}_{\text{E}}\mid \mathbf{V})= (\text{B}(\mathbf{V}) \circ p)(\mathbf{u})$. It ensures that the position and orientation are within the same numerical values for estimating the density and the bijection emphasizes our ignorance about interesting regions of space for grasping.
\par
\begin{figure}[h]
    \centering
    \resizebox{\linewidth}{!}{
    \includegraphics[width=.3\textwidth]{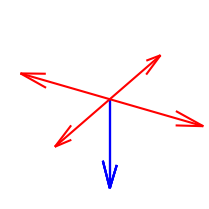}
    \includegraphics[width=.3\textwidth]{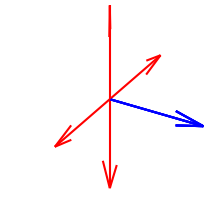}
    \includegraphics[width=.3\textwidth]{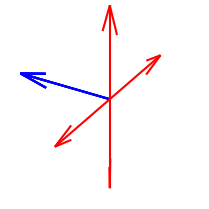}
    \includegraphics[width=.3\textwidth]{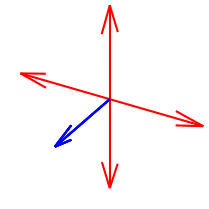}
    \includegraphics[width=.3\textwidth]{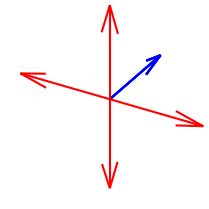}}
    \caption{The modes of the orientation distribution. (left) Encode a top-down approach. (others) Encode side approach.}
    \label{fig:prior_ori_modes}
\end{figure}

\textbf{Orientation} The prior over the orientation $\mathbf{R}_{\text{SE}} \coloneqq \mathbf{q}_{\text{E}}$ is defined as a mixture of \textit{power-spherical} (PS) distributions~\cite{de2020power} with 20 modes $\mathbf{\nu}_{i}$ (Fig.~\ref{fig:prior_ori_modes}). Each mode is itself a mixture that satisfies $p(\mathbf{q}_{\text{E}};\cdot)=p(-\mathbf{q}_{\text{E}};\cdot)$. In total, we have
\begin{equation}p(\mathbf{q}_{\text{E}}) = \frac{1}{20} \sum_{i=1}^{20}\frac{\text{PS}(\mathbf{q}_{\text{E}}; \mathbf{\nu}_{i}, \kappa)}{2} + \frac{\text{PS}(\mathbf{q}_{\text{E}}; -\mathbf{\nu}_{i}, \kappa)}{2}.
\end{equation}
This prior encodes a top-down approach as well as side approaches by its 5 main modes $\nu_{i}$. The 4 additional modes, rotated by $\frac{\pi}{2}$, allows us to explore various orientations. We set the concentration factor $\kappa=8$ for all modes, which keeps the prior gradients low and not hightly regularizes the MAP.
In this way, our prior covers a large part of the rotation space and is sufficiently informative by contrast to a uniform prior over the unit sphere $\mathbb{S}^{3}$.
\par
Finally, $p(\mathbf{h}\mid \mathbf{V}) = p(\mathbf{x}_{\text{E}}\mid \mathbf{V}) p(\mathbf{q}_{\text{E}})$.
\subsection{Density ratio estimation}
The likelihood function $p(S\mid\mathbf{h}, \mathbf{V})$ and the evidence $p(S\mid\mathbf{V})$ are both intractable, which makes standard Bayesian inference procedures such as Markov chain Monte Carlo unusable. 
However, drawing samples from forward models remains feasible with physical simulators, hence enabling likelihood-free Bayesian inference algorithms. 

First, we express the likelihood-to-evidence ratio as,
\begin{align}
    r(S \mid \mathbf{h}, \mathbf{V}) = \frac{p(S \mid \mathbf{h},  \mathbf{V})}{p(S \mid \mathbf{V})} = \frac{p(S, \mathbf{h} \mid \mathbf{V})}{p(S\mid \mathbf{V})p(\mathbf{h} \mid \mathbf{V})}.
    \label{eq:ratio_decomposition}
\end{align}
By adapting the approach described in~\cite{pmlr-v119-hermans20a} for likelihood ratio estimation, we train a neural network classifier $d_\phi$ that we will use to approximate $r(S|\mathbf{h}, \mathbf{V})$.
The network $d_\phi$ is trained to distinguish positive tuples $(S, \mathbf{h}, \mathbf{V})$ (labeled $y=1$) sampled from the joint distribution $p(S, \mathbf{h} \mid \mathbf{V})$ against negative tuples (labeled $y=0$) sampled from the product of marginals $p(S\mid \mathbf{V})p(\mathbf{h} \mid \mathbf{V})$. The Bayes optimal classifier $d^{*}(S,\mathbf{h}, \mathbf{V})$ that minimizes the cross-entropy loss is given by
\begin{equation}
\label{eq:discriminator}
d^{*}(S, \mathbf{h}, \mathbf{V}) = \frac{p(S, \mathbf{h} \mid \mathbf{V})}{p(S \mid \mathbf{V})p(\mathbf{h} \mid \mathbf{V})+ p(S, \mathbf{h} \mid \mathbf{V})},
\end{equation}
which recovers the likelihood ratio $r(S|\mathbf{h})$ as
\begin{equation}
    \label{eq:d_to_r}
    \begin{split}
    \frac{d^{*}(S,\mathbf{h}, \mathbf{V})}{1-d^{*}(S,\mathbf{h}, \mathbf{V})} & = \frac{p(S, \mathbf{h} \mid \mathbf{V})}{p(S \mid \mathbf{V})p(\mathbf{h} \mid \mathbf{V})} = \frac{p(S|\mathbf{h}, \mathbf{V})}{p(S \mid \mathbf{V})}.
    \end{split}
\end{equation}
Therefore, by modelling the classifier with a neural network $d_\phi$ trained on the binary classification problem, we obtain an approximate but amortized and differentiable likelihood ratio  
\begin{equation}
    \hat{r}(S\mid\mathbf{h}, \mathbf{V}) = \frac{d_\phi(S,\mathbf{h}, \mathbf{V})}{1-d_\phi(S,\mathbf{h}, \mathbf{V})}.
\end{equation}
Finally,  the likelihood ratio is combined with the prior to approximate the posterior as
\begin{equation}
\hat{p}(\mathbf{h}|S=1, \mathbf{V}) =  \hat{r}(S=1 \mid  \mathbf{h}, \mathbf{V})p(\mathbf{h} \mid \mathbf{V}),
\end{equation}
which enables immediate posterior inference despite the initial intractability of the likelihood function $p(S \mid \mathbf{h}, \mathbf{V})$ and of the evidence $p(S \mid \mathbf{V})$.

Ensembles tend to produce more conservative posteriors~\cite{hermans2021averting}. In our case, we take 4 models and compute the ratio as
\begin{equation}
    \label{eq:ensembles}
    \log\hat{r} = \log\frac{1}{4}\Sigma_{i=1}^{4}\exp\log\hat{r}_{i}
\end{equation}

The neural network classifiers $d_\phi$ is architectured as follows. 
The hand configuration $\mathbf{h}$ enters the neural network as a tuple of  $(\text{N}_{\text{B}}\times3, \text{N}_{\text{B}}\times4)$ vector where $\text{N}_{\text{B}}$ is the batch size. 
The position is rescaled into a cube of $[-1, 1]$ thanks to a bijection.
In $d_\phi$, $\mathbf{V}$ is fed to a 3D convolutional network made of four convolutional layers followed by a fully connected layer, as in \cite{breyer2020volumetric}, and which goal is to produce a vector embedding of the voxel grid. The voxel embedding, the 4D pose (position and 2D rotation) of the object point cloud $\mathbf{p} = f(\mathbf{V})$ obtained via the TSDF, $S$ and $\mathbf{h}$ are then fed to a subsequent network made of 2 fully connected layers of 256 neurons.
The parameters $\phi$ are optimized using Adam as optimizer.

\subsection{Maximum a posteriori estimation}
\label{subsec:optimization}

Due to the intractability of the likelihood function and of the evidence,
Eq.~(\ref{eq:map}) cannot be solved analytically nor numerically. 
We  rely instead on the  approximation given by the likelihood-to-evidence ratio $\hat{r}$ to find an approximation of the maximum a posteriori (MAP) estimate as
\begin{align}
    \hat{\mathbf{h}}^{*} &= \argmax_{\mathbf{h}}\hat{r}(S=1 \mid \mathbf{h}, \mathbf{V})p(\mathbf{h} \mid \mathbf{V}) \\
    &= \argmin_{\mathbf{h}} -\log \hat{r}(S=1\mid \mathbf{h}, \mathbf{V})p(\mathbf{h}\mid \mathbf{V})
    \label{eq:approximate_map},
\end{align}
which we solve using gradient descent. 
The gradient of Eq.~(\ref{eq:approximate_map}) decomposes as
\begin{equation}
\begin{aligned}
\label{eq:euclidean_grad}
    -\nabla_{(\mathbf{x},\mathbf{q})}\log \hat{r}(S \mid \mathbf{h},\mathbf{V})p(\mathbf{h}\mid \mathbf{V}) = & -\nabla_{(\mathbf{x},\mathbf{q})}\log \hat{r}(S \mid \mathbf{h}, \mathbf{V}) \\ 
    &- \nabla_{(\mathbf{x},\mathbf{q})}\log p(\mathbf{h}\mid \mathbf{V}).
\end{aligned}
\end{equation}
Our prior $p(\mathbf{h}\mid \mathbf{V})$ has analytical gradients. In fact, uniform distributions are set to have null gradient everywhere in the domain. Therefore, $\nabla_{\mathbf{x}}p(\mathbf{h}) = \mathbf{0}$. By contrast, $p(\mathbf{q}_{\text{E}})$ is a weakly informative prior and has a non null gradient from the power spherical distribution. Its derivative with respect to $\mathbf{q}$ is
\begin{equation}
\label{eq:grad_power_spherical}
\begin{split}
\nabla_{\mathbf{q}}p(\mathbf{q};\nu, \kappa) &=  C(\kappa)\kappa(1+\nu^{T}\mathbf{q})^{\kappa-1}\nabla_{\mathbf{q}}(1+\nu^{T}\mathbf{q})\\
 &= C(\kappa)\kappa\mathbf{\nu}(1+\nu^{T}\mathbf{q})^{\kappa-1},
\end{split}
\end{equation}
where $C(\kappa)$ is the normalization term.
Since the likelihood-to-evidence ratio estimator $\hat{r}$ is modelled by a neural network, it is fully differentiable with respect to its inputs and its gradients can be computed by automatic differentiation. 
However, not all variables of the problem are Euclidean variables and naively performing gradient descent would violate our geometric assumptions. 
Let us consider a variable $\mathcal{Z}$ on the smooth Riemannian manifold $\mathcal{M}=\mathbb{R}^{3} \times \mathbb{S}^{3}$  with tangent space $\mathcal{T}_{\mathcal{Z}}\mathcal{M}$ and a function $f : \mathcal{M} \rightarrow \mathbb{R}$. Since $\mathbb{S}^{3}$ is embedded in $\mathbb{R}^{4}$, $f$ can be evaluated on $\mathbb{R}^{3} \times \mathbb{R}^{4} $, leading to the definition of the Euclidean gradients $\nabla f(\mathcal{Z}) \in \mathbb{R}^{3} \times \mathbb{R}^{4}$. In turn, these Euclidean gradients can be transformed into their Riemannian counterparts $\text{grad}f(\mathcal{Z})$ via orthogonal projection $\mathbf{P}_{\mathcal{Z}}$ into the tangent space $\mathcal{T}_{\mathcal{Z}}\mathcal{M}$. Therefore,
\begin{equation}
        \text{grad} f(\mathcal{Z}) =  \mathbf{P}_{\mathcal{Z}}(\nabla f(\mathcal{Z}))
\end{equation}
where the orthogonal projection onto $\mathbb{R}^{3}$ is the identity $\mathbb{I}_{3}$ and the orthogonal projection onto $\mathbb{S}^{3}$ is $\mathbf{P}_{\xi}(\nabla f) = (\mathbb{I}_{4}-\xi\xi^{T})\nabla f$ at $\xi \in \mathbb{S}^{3}$. Thus, we can solve Eq.~(\ref{eq:approximate_map}) by projecting Euclidean gradients of Eq.~(\ref{eq:euclidean_grad}) to the tangent space $\mathcal{T}_{\mathcal{Z}}\mathcal{M}$ and use it in the following update rule
\begin{equation}
    \mathbf{h}_{k+1} = \exp_{\mathbf{h}_{k}}(-\alpha_{k}\text{grad}f(\mathbf{h}_{k}))
\end{equation}
with $\exp_{x}(v):\mathcal{T}_{x}\mathcal{M} \rightarrow \mathcal{M}$ is the exponential map. 

\section{Experiments}
To validate our approach, we perform a series of experiments in simulation as well as in the real setup. We evaluate the performance of our method and determine the transfer capabilities of our network without any fine-tuning.

\subsection{Data generation}\label{sec:data-generation}
The data generating procedure is defined as follow:
\begin{align}
    \label{eq:sample_latent}
    \mathbf{z} &\sim p(\mathbf{z})  \\
    \label{eq:sample_image}
    I_{k} &\sim p(I \mid \mathbf{z}, \mathbf{T}^{k}_{\rm{WC}})  \\
    \label{eq:compute_tsdf}
    \mathbf{V} &= f(\mathcal{I}_{k}, \Gamma_{k})  \\
    \label{eq:sample_hand}
    \{\mathbf{h} & \sim p(\mathbf{h}\mid \mathbf{V})\}  \\
    \label{eq:compute_traj} 
    \{\tau_{1:m} & \sim \Lambda(\tau_{0}, \textsc{ik}(\mathbf{h}), \mathbf{V})\} \\
    \label{eq:sample_success}
    \{S & \sim p(S \mid \tau_{1:m}, \mathbf{z})\}
\end{align}

We use Pybullet~\cite{coumans2020} for implementing these functions. We use the same object assets than VGN~\cite{breyer2020volumetric} for the training and testing. The latent variables $\mathbf{z}$ are described as follow:
\par
\textbf{Object mesh} We sample uniformly an object mesh from an asset of objects.
\par
\textbf{Pose of the table} \pose{ST} We randomize the position $(x, y) \sim \mathcal{N}(0, 0.008)$ and the rotation $q_{\text{T}} = (0., 0., \sin(\frac{\theta_{\text{Table}}}{2}), \cos(\frac{\theta_{\text{Table}}}{2}))$, $\theta_{\text{Table}} \sim \mathcal{U}(-5, 5)$ of the table with respect to \coordinatesystem{S}.
\par
\textbf{Pose of the object} \pose{TO} We randomize the position $(x, y) \sim \mathcal{U}(\frac{-l}{2}, \frac{l}{2})$ and the orientation $q_{\text{O}} = (0., 0., \sin(\frac{\theta_{\text{O}}}{2}), \cos(\frac{\theta_{\text{O}}}{2}))$, $\theta_{\text{O}} \sim \mathcal{U}(0, 2\pi)$ of the object with respect to \coordinatesystem{T}.
\par
\textbf{Torque applied by the fingers} We randomize the final torque applied by the fingers $\uptau \sim \mathcal{U}(35, 40)$.
\par
\textbf{Lateral friction coefficient} We randomize the lateral friction coefficient $\mu \sim \mathcal{U}(1, 2)$.
\par
\textbf{Spinning friction coefficient} We randomize the spinning friction coefficient $\gamma = \eta\mu, \eta \sim \mathcal{N}(0.002, 0.0001)$.
\par
\textbf{Depth images} We add noise to the rendered depth images in simulation using the additive noise model of \cite{jiang2021synergies} with the same parameters.
\subsection{Simulated Experiments}
We evaluate the performance of our method with the success rate (\%). For one round, procedures from~\eqref{eq:sample_latent} to~\eqref{eq:compute_tsdf} are done. We find the MAP or the MLE by sampling 1000 initial hand configurations from the prior and we take the best one. Then, we perform 300 optimization steps with a step size of $0.005$ for the orientation and $0.008$ for the position. Because of the stochastic nature of our MAP estimate, we recompute the MLE/MAP at a maximum of 3 times if the path planner fails to find a valid path. Our method reaches a success rate of nearly $\mathbf{91}\%$ with the MAP, demonstrating the capabilities to adapt to new objects and correctly lift object. Moreover, the MLE performs slightly lower ($87.3\%$) than the MAP. Our weakly informative prior explains the difference in success rates and motivates the use of a Bayesian approach.


\subsection{Real Robot Experiments}
We carry out experiments with a Robotiq 3-finger gripper attached to a UR5 robotic arm, as shown in Fig.~\ref{fig:scene}. A Intel Realsense D435i depth sensor is mounted to the flange of the robotic arm. It produces $848 \times 480$ depth images which are integrated into a TSDF with a resolution of $N=40$ for the network and a resolution of $N=120$ for collision detection using Open3D~\cite{Zhou2018}. The transformation \pose{FC} is calibrated using hand-eye calibration from OpenCV~\cite{opencv_library}. All the devices are handled within the ROS framework.
\begin{figure}
    \centering
    \resizebox{\linewidth}{!}{
    \includegraphics{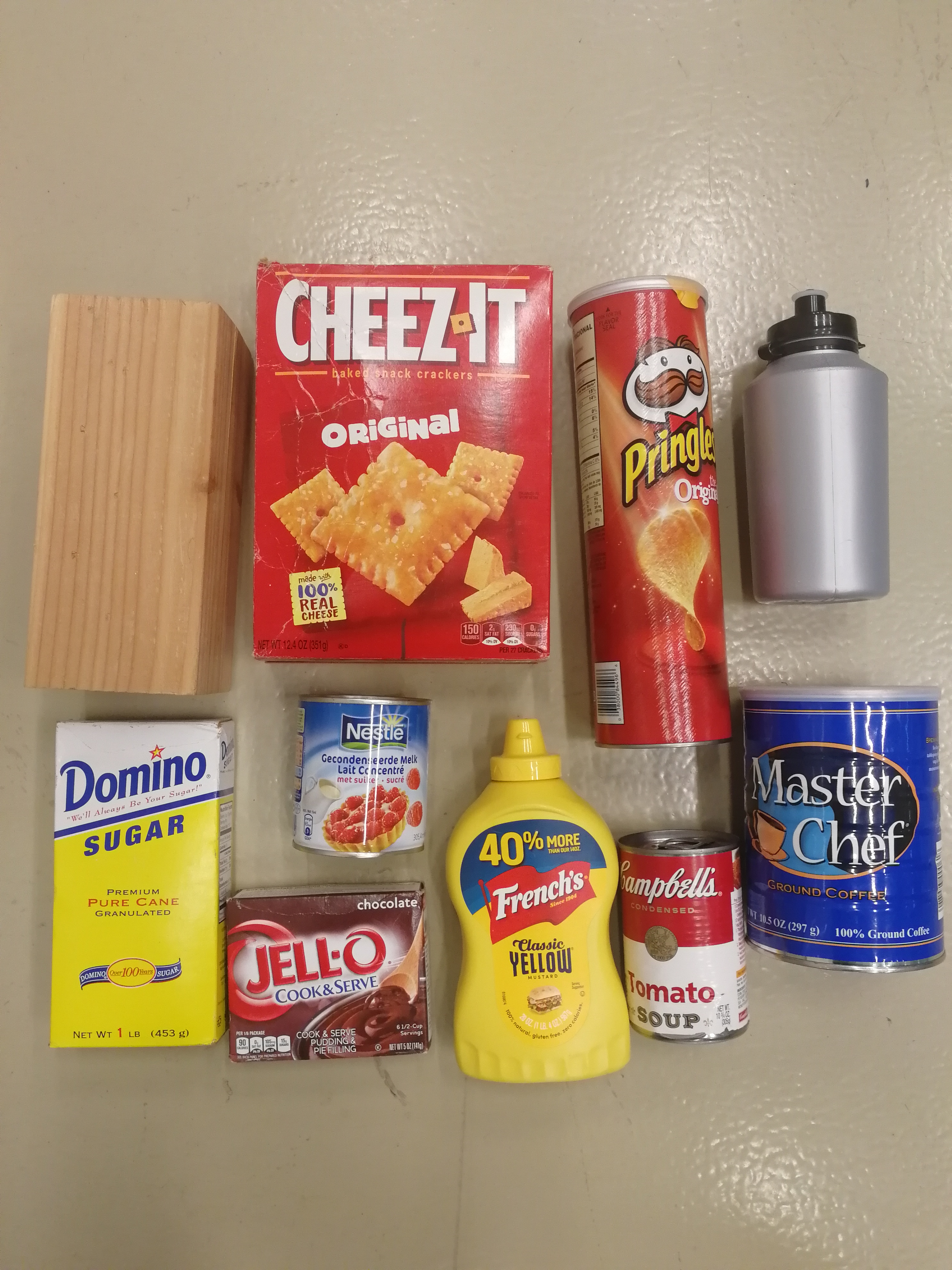}
    \includegraphics{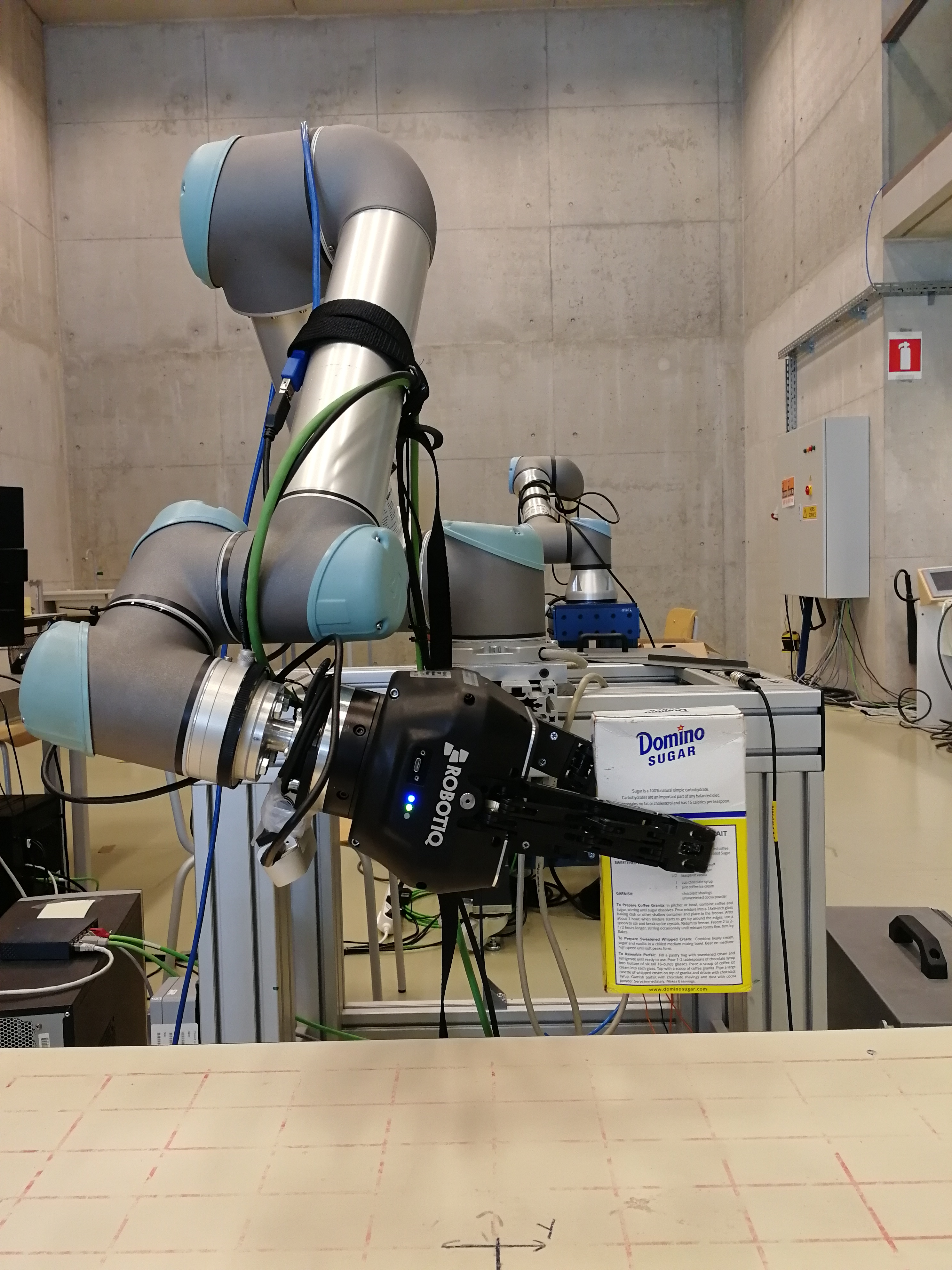}
    }
    \caption{(left) Object assets used in the real setup. (right) Example of side grasp.}
    \label{fig:object_assets}
\end{figure}
We performs 100 rounds with a protocol similar to the simulation experiments. We randomly select 1 object from the 10 test objects and put it randomly on the table by hand. The objects are chosen between seen and unseen objects during training and for their availability in the lab. 
Our success rate of $\mathbf{90}\%$
is similar than in simulation, which indicates that the simulation-to-reality transfer works well. Our approximate ratio learnt successfully several modes to grasp an object and can switch most of the time between them if the path planner fails (Fig~.\ref{fig:object_assets}).

In simulation as well as in the real setup, half of the failure cases are due to the path planner and half are due to wrong hand configurations making the object slip. We leave the improvement of these parts as future work. 

\section{Conclusion}
We demonstrate the usefulness and applicability of simulation-based Bayesian inference to robotic grasping. Our results show promising performance for determining 6 DoF grasp poses.
Nevertheless, our task is rather simple compared to others benchmarks. In the next step, we plan to challenge our method to more complex tasks such as grasping in cluttered environments.

\section*{Acknowledgement}
Norman Marlier would like to acknowledge the Belgian Fund for Research training in Industry and Agriculture for its financial support (FRIA grant). Computational resources have been provided by the Consortium des Équipements de Calcul Intensif (CÉCI), funded by the Fonds de la Recherche Scientifique de Belgique (F.R.S.-FNRS) under Grant No. 2.5020.11 and by the Walloon Region. Gilles Louppe is recipient of the ULiège - NRB Chair on Big Data and is thankful for the support of the NRB

\bibliographystyle{IEEEtran}
\bibliography{IEEEabrv, bibliography.bib}

\end{document}